# Bridging the Gap between Reinforcement Learning and Knowledge Representation: A Logical Off- and On-Policy Framework

Emad Saad [1]

**Abstract.** Knowledge Representation is important issue in reinforcement learning. In this paper, we bridge the gap between reinforcement learning and knowledge representation, by providing a rich knowledge representation framework, based on normal logic programs with answer set semantics, that is capable of solving model-free reinforcement learning problems for more complex domains and exploits the domain-specific knowledge. We prove the correctness of our approach. We show that the complexity of finding an offline and online policy for a model-free reinforcement learning problem in our approach is NP-complete. Moreover, we show that any model-free reinforcement learning problem in MDP environment can be encoded as a SAT problem. The importance of that is model-free reinforcement learning problems can be now solved as SAT problems.

## 1 Introduction

Reinforcement learning is the problem of learning to act by trial and error interaction in dynamic environments. Under the assumption that a complete model of the environment is known, a reinforcement learning problem is modeled as a Markov Decision Process (MDP), in which an optimal policy can be learned. Operation research methods, in particular dynamic programming by value iteration, have been extensively used to learn the optimal policy for a reinforcement learning problem in MDP environment. However, an agent may not know the model of the environment. In addition, an agent may not be able to consider all possibilities and use its knowledge to plan ahead, because of the agent's limited computational abilities to consider all states systematically [4]. Therefore, Q-learning [4] and SARSA [17] are proposed as model-free reinforcement learning algorithms that learn optimal policies without the need for the agent to know the model of the environment.

Q-learning and SARSA are incremental dynamic programming algorithms, that learns optimal policy from actual experience from interaction with the environment, where to guarantee convergence the following assumptions must hold; the action-value function is represented as a look-up table; the environment is a deterministic MDP; for each starting state and action, there are an infinite number of episodes; and the learning rate is decreased appropriately over time. However, these assumptions imply that all actions are tried in every possible state and every state must be visited infinitely many times, which leads to a slow convergence, although, it is sufficient for the agent to try all possible actions in every possible state only once to learn about the reinforcements resulting from executing actions in states. In addition, in some situations it is not possible for the agent to visit a state more than once. Consider a deer that eats in an area where a cheetah appears and the deer flees and survived. If the deer revisits this area again it will be eaten and does not learn anymore. This is unavoidable in Q-learning and SARSA because of the iterative dynamic programming approach they adopt and their convergence assumptions. Moreover, dynamic programming methods use primitive representation of states and actions and do not exploit domain-specific knowledge of the problem domain, in addition they solve MDP with relatively small domain sizes [16]. However, using richer knowledge representation frameworks for MDP allow to efficiently find optimal policies in more complex and larger domains.

[1] Department of Computer Science, Gulf University for Science and Technology, Kuwait, email: saad.e@gust.edu.kw

A logical framework to model-based reinforcement learning has been proposed in [19] that overcomes the representational limitations of dynamic programming methods and capable of representing domain specific knowledge. The framework in [19] is based on the integration of model-based reinforcement learning in MDP environment with normal hybrid probabilistic logic programs with probabilistic answer set semantics [23] that allows representing and reasoning about a variety of fundamental probabilistic reasoning problems including probabilistic planning [18], contingent probabilistic planning [21], the most probable explanation in belief networks, and the most likely trajectory [20].

In this paper we integrate model-free reinforcement learning with normal logic programs with answer set semantics and SAT, providing a logical framework to model-free reinforcement learning using Q-learning and SARSA update rules to learn the optimal off- and on-policy respectively. This framework is considered a model-free extension to the model-based reinforcement learning framework of [19]. The importance of the proposed framework is twofold. First, the proposed framework overcomes the representational limitations of dynamic programming methods to model-free reinforcement learning and capable of representing domain-specific knowledge, and hence bridges the gap between reinforcement learning and knowledge representation. Second, it eliminates the requirement of visiting every state infinitely many times which is required for the convergence of the Q-learning and SARSA.

This integration is achieved by encoding the representation of a model-free reinforcement learning problem in a new high level action language we develop in this paper called, $\mathcal{B}_Q$, into normal logic program with answer set semantics, where all actions are tried in every state only once. We show the correctness of the translation. We prove that the complexity of finding an off- and on-policy in our ap-proach is NP-complete. In addition, we show that any model-free reinforcement learning problem in MDP environment can be encoded as SAT problem.

## 2 Preliminaries

As in the underlying assumptions of the original Q-learning and SARSA, the subsequent results in the rest of this paper assume that the considered MDPs are deterministic. Normal logic programs [7] and Q-learning [4] and SARSA [17] are reviewed in this section.

### 2.1 Normal Logic Programs

Let $\mathcal{L}$ be a first-order language with finitely many predicate symbols, function symbols, constants, and infinitely many variables. The Herbrand base of $\mathcal{L}$ is denoted by $\mathcal{B}$. A Herbrand interpretation is a subset of the Herbrand base $\mathcal{B}$. A normal logic program is a finite set of rules of the form

$$a \leftarrow a_1, \ldots, a_n, not\ b_1, \ldots, not\ b_m$$

Where $a, a_1, \ldots, a_n, b_1, \ldots, b_m$ are atoms and $not$ is the negation-as-failure. A normal logic program is ground if no variables appear in any of its rules. Let $\Pi$ be a ground normal logic program and $S$ be a Herbrand interpretation, then, we say that the above rule is satisfied by $S$ iff $a \in S$, whenever $\{a_1, \ldots, a_n\} \subseteq S$ and $\{b_1, \ldots, b_m\} \cap S = \emptyset$, or for some $i \in \{1, \ldots, n\}$, $j \in \{1, \ldots, m\}$, $a_i \notin S$ or $b_j \in S$.

A Herbrand model of $\Pi$ is a Herbrand interpretation that satisfies every rule in $\Pi$. A Herbrand interpretation $S$ of a normal logic program $\Pi$ is said to be an answer set of $\Pi$ if $S$ is the minimal Herbrand model (with respect to the set inclusion) of the reduct, denoted by $\Pi^S$, of $\Pi$ w.r.t. $S$, where

$$a \leftarrow a_1, \ldots, a_n \in \Pi^S \text{ iff}$$
$$a \leftarrow a_1, \ldots, a_n, not\ b_1, \ldots, not\ b_m \in \Pi$$

and $\{b_1, \ldots, b_m\} \cap S = \emptyset$

### 2.2 Q-learning and SARSA

Q-learning learns the optimal Q-function, $Q^*$, from the agent's experience (set of episodes) by repeatedly estimating the optimal Q-value for every state-action pair $Q^*(s, a)$. The Q-value, $Q(s, a)$, given a policy (a mapping from states to actions), is defined as the expected sum of discounted rewards resulting from executing the action $a$ in a state $s$ and then following the policy thereafter. Given $Q^*$, an optimal policy, $\pi^*$, can be determined by identifying the optimal action in every state, where $a$ is optimal in a state $s$, i.e., $\pi^*(s) = a$, if $\pi^*(s) = \arg\max_{a'} Q^*(s, a') = a$ and $a'$ is executable in $s$. An episode is an exploration of the environment which is a sequence of state-action-reward-state of the form $e \equiv s_0, a_0, r_1, s_1, a_1, r_2, \ldots, s_{n-1}, a_{n-1}, r_n, s_n$, where for each $(0 \leq t \leq n-1)$, $s_t, a_t, r_{t+1}, s_{t+1}$ means that an agent executed action $a_t$ in state $s_t$ and rests in state $s_{t+1}$ where it received reward $r_{t+1}$. $s_0$ denotes an initial state and $s_n$ is a terminal (goal) state. Given that the agent sufficiently explored the environment, the optimal Q-values are repeatedly estimated by the following algorithm:

$\forall (s, a)$ initialize $Q(s, a)$ arbitrary
**Repeat** forever for each episode

  Select the initial state $s_t$ of an episode

  **Repeat**

  Choose an action $a_t$ for the current state $s_t$
  Execute the action $a_t$ in $s_t$
  Observe the subsequent state $s_{t+1}$
  Receive an immediate reward $\mathcal{R}(s_t, a_t, s_{t+1})$
  $Q(s_t, a_t) \leftarrow (1 - \alpha)Q(s_t, a_t) + \alpha[\mathcal{R}(s_t, a_t, s_{t+1})$
  $+ \gamma \max_a Q(s_{t+1}, a)]$
  Set $s_t \leftarrow s_{t+1}$
  **Until** $s_t$ is the end of an episode

where $\alpha$ is the learning rate, $\gamma$ is the discount factor, and $\mathcal{R}(s_t, a_t, s_{t+1})$ is the reward received in $s_{t+1}$ from executing $a_t$ in $s_t$. Q-learning is an offline algorithm that learns the optimal Q-function while executing another policy. Under the same convergence assumptions as in Q-learning, SARSA [17] has been developed as an online model-free reinforcement learning algorithm, that learns optimal Q-function while exploring the environment. Similar to Q-learning, SARSA is an iterative dynamic programming algorithm whose update rule is given by:

$$Q(s_t, a_t) \leftarrow (1-\alpha)Q(s_t, a_t) + \alpha[\mathcal{R}(s_t, a_t, s_{t+1}) + Q(s_{t+1}, a_{t+1})]$$

In addition, SARSA converges slowly to $\mathsf{Q}^*$, since it requires every state to be visited infinitely many times with all actions are tried. Although, it is sufficient for an agent to try all possible actions in every possible state only once to learn about the reinforcements resulting from executing every possible action in every possible state. This assumption could not be eliminated in Q-learning and SARSA, since both are iterative dynamic programming algorithms. However, under the assumption that the environment is finite-horizon Markov decision process with finite length episodes, estimating the optimal Q-function, $Q^*$ for Q-learning, can be simply computed recursively as:

$$Q^*(s_t, a_t) = \mathcal{R}(s_t, a_t, s_{t+1}) + \gamma \max_a Q^*(s_{t+1}, a)$$
$$= \max_a (\mathcal{R}(s_t, a_t, s_{t+1}) + \gamma\ Q^*(s_{t+1}, a))$$
$$= \max_{e \in \mathbf{E}} \left[ \sum_{i=t}^{n-1} \gamma^i \mathcal{R}(s_i, a_i, s_{i+1}) \right] \quad (1)$$

Similarly, the estimate of the optimal Q-function for SARSA can be described as:

$$Q^*(s_t, a_t) = \mathcal{R}(s_t, a_t, s_{t+1}) + \gamma Q^*(s_{t+1}, a_{t+1})$$
$$= \sum_{i=t}^{n-1} \gamma^i \mathcal{R}(s_i, a_i, s_{i+1}) \quad (2)$$

Equations (1) and (2) show that it is sufficient to consider the rewards collected from the set of all episodes, $\mathbf{E}$, only once to calculate estimate of the optimal Q-function, $Q^*$, which eliminates the need to visit every possible state infinitely many times.

Unlike Q-learning, our estimate of $Q^*$, can be computed online as well as offline. It can be computed online by accumulating estimate of $Q^*$ during the exploration of the environment. On the other hand, it can be computed offline by first exploring the environment and collecting the set of all possible episodes, then computing estimate of $Q^*$.

## 3 Action Language $\mathcal{B}_\mathcal{Q}$

This section develops the syntax and semantics of the action language, $\mathcal{B}_\mathcal{Q}$, that allows the representation of model-free reinforcement learning problems, which extends the action language $\mathcal{B}$ [8].

### 3.1 Language syntax

A fluent is a predicate, which may contain variables, that describes a property of the environment. Let $\mathcal{F}$ be a set of fluents and $\mathcal{A}$ be a set of action names that can contain variables. A fluent literal is either a fluent $f \in \mathcal{F}$ or $\neg f$, the negation of $f$. Conjunctive fluent formula is a conjunction of fluent literals of the form $l_1 \wedge \ldots \wedge l_n$, where $l_1, \ldots, l_n$ are fluent literals. Sometimes we abuse the notation and refer to a conjunctive fluent formula as a set of fluent literals ($\emptyset$ denotes $true$). An action theory, $\mathbf{T}$, in $\mathcal{B}_\mathcal{Q}$ is a tuple of the form $\mathbf{T} = \langle S_0, \mathcal{D}, \gamma \rangle$, where $S_0$ is a proposition of the form (3), $\mathcal{D}$ is a set of propositions from (4-6), and $0 \leq \gamma < 1$ is a discount factor as follows:

$$\mathbf{initially}\{\psi_i | 1 \leq i \leq n\} \quad (3)$$
$$\mathbf{executable}\ a\ \mathbf{if}\ \psi \quad (4)$$
$$l\ \mathbf{if}\ \psi \quad (5)$$
$$a\ \mathbf{causes}\ \phi : r\ \mathbf{if}\ \psi \quad (6)$$

where $l$ is a fluent literal, $\phi, \psi, \psi_1, \ldots, \psi_n$ are conjunctive fluent formulas, $a \in \mathcal{A}$ is an action, and $r$ is a real number in $\mathbb{R}$.

Proposition (3) represents the set of possible initial states. Proposition (4) states that an action $a$ is executable in any state in which $\psi$ holds, where each variable that appears in $a$ also appears in $\psi$. *Indirect effect of action* is described by proposition (5), which says that $l$ holds in every state in which $\psi$ also holds. A proposition of the form (6) represents the conditional effects of an action $a$ along with the rewards received in a state resulting from executing $a$. All variables that appear in $\phi$ also appear in $a$ and $\psi$. Proposition (6) says that $a$ causes $\phi$ to hold with reward $r$ is received in a successor state to a state in which $a$ is executed and $\psi$ holds. An action theory is ground if it does not contain any variables.

**Example 1** *Consider an elevator of n-story building domain adapted from [5] that is represented by an action theory, $\mathbf{T} = \langle S_0, \mathcal{D}, \gamma \rangle$, in $\mathcal{B}_\mathcal{Q}$, where $S_0$ is described by (7) (j is a particular value in $\{1, \ldots, n\}$ and $1 \leq i \leq k$ for $k \leq n$) and $\mathcal{D}$ is represented by (8)-(14).*

$$\mathbf{initially}\{\{on(i), \neg opened, current(j)\}\} \quad (1)$$
$$up(N)\ \mathbf{causes}\ current(N), \neg on(N), opened : r\ \mathbf{if}$$
$$on(N), \neg opened \quad (2)$$
$$down(N)\ \mathbf{causes}\ current(N), \neg on(N), opened : r\ \mathbf{if}$$
$$on(N), \neg opened \quad (3)$$
$$close\ \mathbf{causes}\ \neg opened : r\ \mathbf{if}\ opened \quad (4)$$
$$current(N)\ \mathbf{if}\ \neg current(M), N \neq M \quad (5)$$
$$\mathbf{executable}\ up(N)\ \mathbf{if}\ current(M), M < N \quad (6)$$
$$\mathbf{executable}\ down(N)\ \mathbf{if}\ current(M), M > N \quad (7)$$
$$\mathbf{executable}\ close\ \mathbf{if}\ \{\} \quad (8)$$

*The actions in the elevator domain are $up(N)$ for move up to floor $N$, $down(N)$ for move down to floor $N$, and $close$ for closing the elevator door. The predicates $current(N)$, $on(N)$, and $opened$ are fluents represent respectively that the elevator current floor is $N$, light of floor $N$ is on, and elevator door is opened. The target is to get all floors serviced and $\neg on(N)$ is true for all $N$.*

### 3.2 Semantics

We say a set of ground literals $\phi$ is consistent if it does not contain a pair of complementary literals. If a literal $l \in \phi$, then we say $l$ holds in $\phi$, and $l$ does not hold in $\phi$ if $\neg l \in \phi$. A set of literals $\sigma$ holds in $\phi$ if $\sigma$ is contained in $\phi$, otherwise, $\sigma$ does not hold in $\phi$. We say that a set of literals $\phi$ satisfies an indirect effect of action of the form (5), if $l$ belongs to $\phi$ whenever $\psi$ is contained in $\phi$ or $\psi$ is not contained in $\phi$. Let $\mathbf{T}$ be an action theory in $\mathcal{B}_\mathcal{Q}$ and $\phi$ be a set of literals. Then $\mathcal{C}_\mathbf{T}(\phi)$ is the smallest set of literals that contains $\phi$ and satisfies all indirect effects of actions propositions in $\mathbf{T}$. A state $s$ is a complete and consistent set of literals that satisfies all the indirect effects of actions propositions in $\mathbf{T}$.

**Definition 1** *Let $\mathbf{T} = \langle S_0, \mathcal{D}, \gamma \rangle$ be a ground action theory in $\mathcal{B}_\mathcal{Q}$, $s$ be a state, $a$ causes $\phi : r$ if $\psi$ be a proposition in $\mathcal{D}$. Then, $s' = \mathcal{C}_\mathbf{T}(\Phi(a, s))$ is the state resulting from executing $a$ in $s$, given that $a$ is executable in $s$, where $\Phi(a, s)$ is defined as:*

- $l \in \Phi(a, s)$ and $\neg l \notin \Phi(a, s)$ if $l \in \phi$ and $\psi \subseteq s$.
- $\neg l \in \Phi(a, s)$ and $l \notin \Phi(a, s)$ if $\neg l \in \phi$ and $\psi \subseteq s$.
- Otherwise, $l \in \Phi(a, s)$ iff $l \in s$ and $\neg l \in \Phi(a, s)$ iff $\neg l \in s$.

*where the reward received in $s'$ is $\mathcal{R}(s, a, s') = r$.*

An episode in $\mathbf{T} = \langle S_0, \mathcal{D}, \gamma \rangle$ is an expression of the form $e \equiv s_0, a_0, r_1, s_1, a_1, \ldots, s_{n-1}, a_{n-1}, r_n, s_n$, where for each $(0 \leq t \leq n-1)$, $s_{t+1} = \mathcal{C}_\mathbf{T}(\Phi(a_t, s_t))$ and $\mathcal{R}(s_t, a_t, s_{t+1}) = r_{t+1}$.

**Definition 2** *Let $\mathbf{T} = \langle S_0, \mathcal{D}, \gamma \rangle$ be a ground action theory and $\mathbf{E}$ be the set of all episodes in $\mathbf{T}$. Then, for $(0 \leq t \leq n-1)$, where $Q^*(s_n, a_n) = 0$, the optimal Q-function, $Q^*$, for Q-learning and SARSA are respectively estimated by*

$$Q^*(s_t, a_t) = \max_{e \in \mathbf{E}} \sum_{i=t}^{n-1} \gamma^i \mathcal{R}(s_i, a_i, s_{i+1})$$
$$Q^*(s_t, a_t) = \sum_{i=t}^{n-1} \gamma^i \mathcal{R}(s_i, a_i, s_{i+1})$$

Considering SARSA, the optimal Q-function can be computed incrementally as follows. For any episode in $\mathbf{E}$, the optimal Q-value for the initial state-action pair is estimated by

$$Q^*(s_0, a_0) = \sum_{t=0}^{n-1} \gamma^t \mathcal{R}(s_t, a_t, s_{t+1})$$

that is calculated online during the exploration of the environment. Then, for any state-action pair, $(s_t, a_t)$, in the episode, $Q^*(s_t, a_t)$, is calculated from $Q^*(s_0, a_0)$ by

$$Q^*(s_t, a_t) = \frac{Q^*(s_0, a_0) - \sum_{i=1}^{t} \gamma^{i-1} \mathcal{R}(s_{i-1}, a_{i-1}, s_i)}{\gamma^t} \quad (15)$$

However, for Q-learning, $Q^*$ can be computed incrementally as well by first computing $Q^*$ incrementally using (15), then (16) is used as an update rule only once, where for $0 \leq t < n - 1$

$$Q^*(s_t, a_t) = \mathcal{R}(s_t, a_t, s_{t+1}) + \gamma \max_a Q^*(s_{t+1}, a) \quad (16)$$

Notice that, unlike [4], by using (15) and (16), Q-learning can be computed online during the exploration of the environment as well as offline.

## 4 Off- and On-Policy Model-free Reinforcement Learning Using Answer Set Programming

We provide a translation from any action theory $\mathbf{T} = \langle S_0, \mathcal{D}, \gamma \rangle$, a representation of a model-free reinforcement learning problem into a normal logic program with answer set semantics $\Pi_\mathbf{T}$, where the rules in $\Pi_\mathbf{T}$ encode (1) the set of possible initial states $S_0$, (2) the transition function $\Phi$, (3) the set of propositions in $\mathcal{D}$, (4) and the discount factor $\gamma$. The answer sets of $\Pi_\mathbf{T}$ correspond to episodes in $\mathbf{T}$, with associated estimated optimal Q-values. This translation follows some related translations described in [24, 18, 19].

We assume the environment is a finite-horizon Markov decision process, where the length of each episode is known and finite. We use the predicates; $holds(L, T)$ to represent a literal $L$ holds at time moment $T$; $occ(A, T)$ for action $A$ executes at time $T$; $reward(r, a, T)$ for reward received at time $T$ after executing $a$ is $r$; $Q(V, A, T)$ says the estimate of the optimal Q-value of the initial state-action pair, in a given episode, $T$ steps from the initial state is $V$; and $factor(\gamma)$ for the discount factor. We use lower case letters to represent constants and upper case letters to represent variables.

Let $\Pi_\mathbf{T}$ be the normal logic program translation of $\mathbf{T} = \langle S_0, \mathcal{D}, \gamma \rangle$ that contains a set of rules described as follows. To simplify the presentation, given $p$ is a predicate and $\psi = \{l_1, \ldots, l_n\}$ be a set of literals, we use $p(\psi)$ to denote $p(l_1), \ldots, p(l_n)$.

● For each action $a \in \mathcal{A}$, $\Pi_\mathbf{T}$ includes the set of facts

$$action(a) \leftarrow \quad (17)$$

● Literals describe states of the world are encoded by

$$literal(A) \leftarrow atom(A) \quad (18)$$
$$literal(\neg A) \leftarrow atom(A) \quad (19)$$

where $atom(A)$ is a set of facts that describe the properties of the world. To specify that $A$ and $\neg A$ are contrary literals the following rules are added to $\Pi_\mathbf{T}$.

$$contrary(A, \neg A) \leftarrow atom(A) \quad (20)$$
$$contrary(\neg A, A) \leftarrow atom(A) \quad (21)$$

● The set of initial states, $\mathbf{initially}\{\psi_i, 1 \leq i \leq n\}$, is encoded as follows. Let $s_1, s_2, \ldots, s_n$ be the set of initial states, where for $1 \leq i \leq n$, $s_i = \{l_1^i, \ldots, l_m^i\}$. Moreover, let $s = s_1 \cup s_2 \cup \ldots \cup s_n$, $s' = s_1 \cap s_2 \cap \ldots \cap s_n$, $\widehat{s} = s - s'$, and $s'' = \{l \mid l \in \widehat{s} \vee \neg l \in \widehat{s}\}$. Intuitively, for any literal $l$ in $\widehat{s}$, if $l$ or $\neg l$ belongs to $\widehat{s}$, then $s''$ contains only $l$. For each literal $l \in s'$, $\Pi_\mathbf{T}$ includes

$$holds(l, 0) \leftarrow \quad (22)$$

which says $l$ holds at time 0. Literals in $s'$ belong to every initial state. For each $l \in s''$, $\Pi_\mathbf{T}$ includes

$$holds(l, 0) \leftarrow not\ holds(\neg l, 0) \quad (23)$$
$$holds(\neg l, 0) \leftarrow not\ holds(l, 0) \quad (24)$$

which says that $l$ (similarly $\neg l$) holds at time 0, if $\neg l$ (similarly $l$) does not hold at the time 0.

● Each proposition of the form (4) is encoded in $\Pi_\mathbf{T}$ as

$$exec(a, T) \leftarrow holds(\psi, T) \quad (25)$$

● Each $a$ **causes** $\phi$ : $r$ **if** $\psi$ in $\mathcal{D}$ is encoded as

$$holds(l_i, T+1) \leftarrow occ(a, T), exec(a, T), holds(\psi, T) \quad (26)$$

$\forall l_i \in \phi$ and $\phi = \{l_1, \ldots, l_m\}$, which says that if $a$ occurs at time $T$ and $\psi$ holds at the same time moment, then $l_i$ holds at time $T+1$.

● The reward $r$ received at time $T+1$ after executing $a$ at time $T$ given that $a$ is executable is encoded by

$$reward(r, a, T+1) \leftarrow occ(a, T), exec(a, T) \quad (27)$$

● Estimate of the optimal Q-value of an initial state-action pair, in a given episode, $T+1$ steps away from the initial state, is equal to the estimate of the optimal Q-value of the same initial state-action pair, in the same episode, $T$ steps away from the initial state added to the discounted reward (by $\gamma^T$) received at time $T+1$, where $V \in \mathbb{R}$ and $0 \leq \gamma < 1$.

$$Q(V + r * \gamma^T, a, T+1) \leftarrow Q(V, a', T), factor(\gamma),$$
$$reward(r, a, T+1),$$
$$occ(a, T), exec(a, T), holds(\psi, T), holds(\phi, T+1) \quad (28)$$

● The following rule says that $L$ holds at the time moment $T+1$ if it holds at the time moment $T$ and its contrary does not hold at the time moment $T+1$.

$$holds(L, T+1) \leftarrow holds(L, T), not\ holds(L', T+1),$$
$$contrary(L, L') \quad (29)$$

● A literal $L$ and its negation $\neg L$ cannot hold at the same time is encoded in $\Pi_\mathbf{T}$ by

$$\leftarrow holds(L, T), holds(\neg L, T) \quad (30)$$

● Rules that generate actions occurrences once at a time are encoded by

$$occ(AC^i, T) \leftarrow action(AC^i), not\ abocc(AC^i, T) \quad (31)$$
$$abocc(AC^i, T) \leftarrow action(AC^i), action(AC^j), occ(AC^j, T),$$
$$AC^i \neq AC^j \quad (32)$$

● Let $\mathcal{G} = g_1 \wedge \ldots \wedge g_m$ be a goal expression, then $\mathcal{G}$ is encoded in $\Pi_\mathbf{T}$ as

$$goal \leftarrow holds(g_1, T), \ldots, holds(g_m, T) \quad (33)$$

Estimates of the optimal Q-value of initial state-action pair, $Q^*(s_0, a_0)$, is represented in $\Pi_\mathbf{T}$ by $Q(V, A, T)$, for $0 \leq T \leq n$, where $Q(V, A, n)$ represents the estimate of $Q^*(s_0, a_0)$ at the end of episode of length $n$. These Q-values, $Q(V, A, n)$, can be computed online during the exploration of the environment as well as offline after the exploration of the environment. Moreover, the action generation rules (31) and (32) in our translation, choose actions greedily at random. However, other action selection strategies can be encoded instead.

**Example 2** *The normal logic program encoding, $\Pi_\mathbf{T}$, of the elevator domain described in Example 1 is given as follows, where $\Pi_\mathbf{T}$ consists of the following rules, along with the rules (18), (19), (20), (21), (29), (30), (31), (32):*

$$action(open(N)) \leftarrow$$
$$action(down(N)) \leftarrow$$
$$action(close) \leftarrow$$

*for $1 < N < n$. The atoms $on(.)$, $current(.)$, and $opened$ describe properties of the world that for $1 \leq N \leq n$. are encoded as*

$$atom(on(N)) \leftarrow$$
$$atom(current(N)) \leftarrow$$
$$atom(opened) \leftarrow$$

The initial state is encoded as follows, where $1 < X < k$, for $k < n$ and for some $j$ in $\{1, \ldots, n\}$.

$$holds(on(X), 0) \leftarrow$$
$$holds(current(j), 0) \leftarrow$$
$$holds(\neg opened, 0) \leftarrow$$

The executability conditions of actions, for $1 < N, M < n$, are encoded as

$$exec(up(N), T) \leftarrow holds(current(M), T), M < N$$
$$exec(down(N), T) \leftarrow holds(current(M), T), M > N$$
$$exec(close, T) \leftarrow$$

Effects, rewards, and the Q-value of the initial state-action pair resulting after executing the actions $up(N)$ and $down(N)$, for $1 < N < n$, are given by

$$holds(current(N), T+1) \leftarrow occ(AC, T), exec(AC, T),$$
$$holds(on(N), T), holds(\neg opened, T)$$
$$holds(\neg on(N), T+1) \leftarrow occ(AC, T), exec(AC, T),$$
$$holds(on(N), T), holds(\neg opened, T)$$
$$holds(opened, T+1) \leftarrow occ(AC, T), exec(AC, T),$$
$$holds(on(N), T), holds(\neg opened, T)$$

$$reward(r, AC, T+1) \leftarrow occ(AC, T), exec(AC, T)$$

$$Q(V + r * \gamma^T, AC, T+1) \leftarrow Q(V, A, T), factor(\gamma),$$
$$reward(r, AC, T+1),$$
$$occ(AC, T), exec(AC, T), holds(\psi, T), holds(\phi, T+1)$$

where $AC = \{up(N), down(N)\}$, $Q(0, AC, 0)$ is a fact, $\psi = \{on(N), \neg opened\}$, and $\phi = \{current(N), \neg on(N), opened\}$.

Effects of the $close$ action is given by

$$holds(\neg opened, T+1) \leftarrow occ(close, T), exec(close, T),$$
$$holds(opened, T)$$

The reward received after executing $close$ is given by

$$reward(r, close, T+1) \leftarrow occ(close, T), exec(close, T)$$

Q-value of the initial state-action pair is given by the following rule, where $Q(0, close, 0)$ is a fact.

$$Q(V + r * \gamma^T, close, T+1) \leftarrow Q(V, A, T), factor(\gamma),$$
$$reward(r, close, T+1), occ(close, T),$$
$$exec(close, T), holds(opened, T), holds(\neg opened, T+1)$$

The goal is encoded by the following rule for some $k \leq n$

$$goal \leftarrow holds(\neg on(1), T), \ldots, holds(\neg on(k), T)$$

## 5 Correctness

This section shows the correctness of our translation. We prove that the answer sets of the normal logic program translation of an action theory, $\mathbf{T}$ in $\mathcal{B}_\mathcal{Q}$, correspond to episodes in $\mathbf{T} = \langle S_0, \mathcal{D}, \gamma \rangle$, associated with estimates of the optimal Q-values. Moreover, we show that the complexity of finding a policy for $\mathbf{T}$ in our approach is NP-complete. Let the domain of $\mathbf{T}$ be $\{0, \ldots, n\}$. Let $\Phi$ be a transition function associated with $\mathbf{T}$, $s_0$ is an initial state, and $a_0, \ldots, a_{n-1}$ be a set of actions in $\mathcal{A}$. An episode in $\mathbf{T}$ is state-action-reward-state sequence of the form $e \equiv s_0, a_0, r_1, s_1, a_1, r_2, \ldots, s_{n-1}, a_{n-1}, r_n, s_n$, such that $\forall (0 \leq i \leq n-1)$, $s_i, s_{i+1}$ are states, $a_i$ is an action, $s_{i+1} = \mathcal{C}_\mathbf{T}(\Phi(a_i, s_i))$, and $\mathcal{R}(s_i, a_i, s_{i+1}) = r_{i+1}$.

**Theorem 1** Let $\mathbf{T}$ be an action theory representing a model-free reinforcement learning problem in $\mathcal{B}_\mathcal{Q}$. Then, $s_0, a_0, r_1, s_1, a_1, r_2, \ldots, s_{n-1}, a_{n-1}, r_n, s_n$ is an episode in $\mathbf{T}$ iff $occ(a_0, 0), reward(r_1, a_0, 1), \ldots, occ(a_{n-1}, n-1), reward(r_n, a_{n-1}, n)$ is true in an answer set of $\Pi_\mathbf{T}$.

Theorem 1 says that an action theory, $\mathbf{T}$, in $\mathcal{B}_\mathcal{Q}$, can be translated into a normal logic program, $\Pi_\mathbf{T}$, such that an answer set of $\Pi_\mathbf{T}$ is equivalent to an episode in $\mathbf{T}$.

**Theorem 2** Let $\mathbf{T}$ be an action theory in $\mathcal{B}_\mathcal{Q}$, $S$ be an answer set of $\Pi_\mathbf{T}$, and $\mathbf{E}$ be the set of all episodes in $\mathbf{T}$. Let $\mathcal{OCC}$ be a set such that $s_0, a_0, r_1, s_1, a_1, r_2, \ldots, s_{n-1}, a_{n-1}, r_n, s_n \in \mathbf{E}$ iff $occ(a_0, 0), reward(r_1, a_0, 1), \ldots, occ(a_{n-1}, n-1), reward(r_n, a_{n-1}, n) \equiv o \in \mathcal{OCC}$. Then, the estimate of $Q^*(s_0, a_0)$ is given for Q-learning and SARSA respectively by

$$Q^*(s_0, a_0) = \max_{S \models Q(v, a_{n-1}, n) \wedge S \models o \in \mathcal{OCC}} v$$

$$Q^*(s_0, a_0) = v, \text{ for some } S \models Q(v, a_{n-1}, n) \wedge S \models o \in \mathcal{OCC}$$

Theorem 2 asserts that, given an action theory $\mathbf{T}$ and by considering Q-learning update rule, the expected sum of discounted rewards resulting after executing an action $a_0$ in a state $s_0$ and following the optimal policy thereafter, $Q^*(s_0, a_0)$, is equal to the maximum over $v$, appearing in $Q(v, a_{n-1}, n)$ which is satisfied by every answer set $S$ of $\Pi_\mathbf{T}$ for which $o \equiv occ(a_0, 0), reward(r_1, a_0, 1), \ldots, occ(a_{n-1}, n-1), reward(r_n, a_{n-1}, n)$ is also satisfied. However, by considering the update rule of SARSA, $Q^*(s_0, a_0)$ is equal to $v$ in $Q(v, a_{n-1}, n)$ that is satisfied by some answer set of $\Pi_\mathbf{T}$ for which $o$ is also satisfied. For any $a_t$ and $s_t$ in SARSA, $Q^*(s_t, a_t)$ is calculated from $Q^*(s_0, a_0)$ by (15), where $Q^*(s_n, a_n) = 0$ and $\mathcal{R}(s_{i-1}, a_{i-1}, s_i) = r_i$. But, for Q-learning, for any $a_t$ and $s_t$, $Q^*(s_t, a_t)$ is calculated from $Q^*(s_0, a_0)$ by (15), then (16) is used as an update rule only once.

In addition, we show that any model-free reinforcement learning problem in MDP environment can be encoded as SAT problem. Hence, state-of-the-art SAT solvers can be used to solve model-free reinforcement learning problems. Any normal logic program, $\Pi$, can be translated into a SAT problem, $\mathcal{S}$, where the models of $\mathcal{S}$ are equivalent to the answer sets of $\Pi$ [13]. Hence, the normal logic program encoding of a model-free reinforcement learning problem $\mathbf{T}$ can be translated into an equivalent SAT problem, where the models of $\mathcal{S}$ correspond to episodes in $\mathbf{T}$.

**Theorem 3** Let $\mathbf{T}$ be an action theory in $\mathcal{B}_\mathcal{Q}$ and $\Pi_\mathbf{T}$ be the normal logic program encoding of $\mathbf{T}$. Then, the models of the SAT encoding of $\Pi_\mathbf{T}$ are equivalent to valid episodes in $\mathbf{T}$.

For SAT encoding, the optimal Q-function is computed in a similar way as in the normal logic program encoding of model-free reinforcement learning problems. The transformation step from nor-mal logic program encoding of a model-free reinforcement learning problem into SAT can be avoided, by encoding a model-free reinforcement learning problem directly into SAT [22]. The following corollary shows any model-free reinforcement learning problem can be encoded directly as SAT problem.

**Corollary 1** Let $\mathbf{T}$ be an action theory in $\mathcal{B}_\mathcal{Q}$. Then, $\mathbf{T}$ can be directly encoded as a SAT formula $\mathcal{S}$ where the models of $\mathcal{S}$ are equivalent to valid episodes in $\mathbf{T}$.

Normal logic programs with answer set semantics find optimal policies for model-free reinforcement learning problems in finite horizon MDP environments using the flat representation of the problem domains.

The flat representation of reinforcement learning problem domains is the explicit enumeration of world states [14]. Hence, Theorem 5 follows directly from Theorem 4 [14].

**Theorem 4** *The stationary policy existence problem for finite-horizon MDP in the flat representation is NP-complete.*

**Theorem 5** *The policy existence problem for a model-free reinforcement learning problem in MDP environment using normal logic pro-grams with answer set semantics and SAT is NP-complete.*

## 6 Conclusions and Related Work

We described a high level action language called $\mathcal{B}_\mathcal{Q}$ that allows the representation of model-free reinforcement learning problems in MDP environments. In addition, we introduced online and offline logical framework to model-free reinforcement learning by relating model-free reinforcement learning in MDP environment to normal logic programs with answer set semantics and SAT.

The translation from an action theory in $\mathcal{B}_\mathcal{Q}$ into a normal logic program builds on similar translations described in [24, 18, 19]. The literature is rich with action languages that are capable of represent-ing and reasoning about MDPs and actions with probabilistic effects, which include [1, 2, 6, 9, 12]. The main difference between these languages and $\mathcal{B}_\mathcal{Q}$ is that $\mathcal{B}_\mathcal{Q}$ allows the factored characterization of MDP for model-free reinforcement learning.

Many approaches for solving MDP to find the optimal policy for both reinforcement learning and probabilistic planning have been presented. These approaches can be classified into two main categories of approaches; dynamic programming approaches and the search-based approaches (a detailed survey on these approaches can be found in [10, 2]). However, dynamic programming approaches use primitive domain knowledge representation. On the other hand, the search-based approaches mainly rely on search heuristics which have limited knowledge representation capabilities to represent and use domain-specific knowledge.

A logic based approach for solving MDP, for probabilistic plan-ning, has been presented in [15]. The approach of [15] converts MDP specification of a probabilistic planning problem into a stochastic satisfiability problem and solving the stochastic satisfiability problem instead. First-order logic representation of MDP for model-based reinforcement learning has been described in [11] based on first-order logic programs without nonmonotonic negations. Similar to the first-order representation of MDP in [11], $\mathcal{B}_\mathcal{Q}$ allows objects and relations. However, unlike $\mathcal{B}_\mathcal{Q}$, [11] finds policies in the abstract level. A more expressive first-order representation of MDP than [11] has been presented in [3] that is a probabilistic extension to Reiter's situation calculus. Although more expressive, it is more complex than [11]. Unlike the logical model-based reinforcement learning frame-work of [19] that uses normal hybrid probabilistic logic programs to encode model-based reinforcement learning problems, normal logic program with answer set semantics is used to encode model-free reinforcement learning problems.